# The performances of the Chinese and U.S. Large Language Models on the Topic of Chinese Culture

Feiyan Liu[1], Siyan Zhao[2], Chenxun Zhuo[3], Tianming Liu[4], Bao Ge[1,*]

[1]School of Physics and Information Technology, Shaanxi Normal University, Xi'an

[2]School of History and Culture Tourism, Xi'an University, Xi'an
[3]School of Foreign Language, Northwest University, Xi'an
[4]School of Computing, The University of Georgia, Athens 30602, USA

**Abstract**

Cultural backgrounds shape individuals' perspectives and approaches to problem-solving. Since the emergence of GPT-1 in 2018, large language models (LLMs) have undergone rapid development. To date, the world's ten leading LLM developers are primarily based in China and the United States. To examine whether LLMs released by Chinese and U.S. developers exhibit cultural differences in Chinese-language settings, we evaluate their performance on questions about Chinese culture. This study adopts a direct-questioning paradigm to evaluate models such as GPT-5.1, DeepSeek-V3.2, Qwen3-Max, and Gemini2.5Pro. We assess their understanding of traditional Chinese culture, including history, literature, poetry, and related domains. Comparative analyses between LLMs developed in China and the U.S. indicate that Chinese models generally outperform their U.S. counterparts on these tasks. Among U.S.-developed models, Gemini 2.5Pro and GPT-5.1 achieve relatively higher accuracy. The observed performance differences may potentially arise from variations in training data distribution, localization strategies, and the degree of emphasis on Chinese cultural content during model development.

## 1 Introduction

Before the introduction of the Transformer architecture, deep neural networks primarily relied on recurrent neural networks (RNNs) to process sequential data [1]. However, RNNs were constrained in scale and capacity and performed poorly on long text sequences. In 2017, a Google research team proposed the Transformer architecture [2], which provided the theoretical foundation for subsequent large language models (LLMs) and marked a new phase in natural language processing.

Natural language processing (NLP) is a subfield of artificial intelligence whose core objective is to develop algorithms and models capable of understanding, interpreting, and generating human language [3]. NLP tasks span a broad spectrum, including natural language understanding (NLU) tasks such as sentiment analysis, text classification, and machine translation [4]. Before the rise of deep learning, NLP primarily relied on traditional machine learning models. For instance, sentiment analysis typically combined hand-crafted features with classifiers such as logistic regression (LR) and support vector machines (SVM) [5]. However, this paradigm required extensive manual feature engineering, such as part-of-speech tagging, and was complex with limited generalization capability.

Subsequently, Devlin et al. [6] developed BERT, a Transformer-based model pre-trained on large-scale corpora, which can be fine-tuned for numerous downstream tasks such as named entity recognition. Meanwhile, Radford et al. [7] introduced GPT-1, ushering NLP into the era of generative large language models. These models not only support interaction through natural language prompts but also learn richer linguistic structures. The subsequent development of GPT-2 [8] demonstrated that large-scale language models trained on sufficiently diverse datasets can accomplish many tasks via prompts without task-specific fine-tuning. In 2020, Raffel et al. [9] proposed the text-to-text framework, which offers a unified perspective on NLP by casting all tasks as text generation. By this stage, the Transformer architecture had fundamentally reshaped the field of NLP.

Since then, large language models (LLMs) have advanced rapidly, evolving from pure text processors to multimodal systems that can jointly process text and images. Through self-supervised training on vast text corpora, LLMs learn statistical regularities and semantic structures of language and thereby acquire capabilities in generation, comprehension, reasoning, and interaction. LLMs demonstrate strong performance across diverse application domains. In medicine, for example, Raffaele et al. evaluated LLMs on the Royal College of Ophthalmologists' examinations. They reported that an LLM-based chatbot achieved an accuracy of 82.9% on postgraduate ophthalmology specialty exams, exceeding the passing threshold for certification [10]. In social psychology, Yair et al. used LLMs that were not explicitly trained on social norms to classify 25 scenarios designed to assess social emotions. The study showed that the models correctly classified 16 scenarios and achieved their highest accuracy when identifying violations of social norms such as trust and care [11].

Moreover, LLMs also exhibit notable performance in educational and cultural applications. Lü Wei et al. proposed an LLM-driven, multi-agent framework for translating classical Chinese that not only better preserves cultural allusions and enhances semantic consistency but also significantly reduces processing time, thereby providing a new approach to cross-lingual translation [12]. CultureLLM and the semantic augmentation method proposed by Li Cheng et al. outperformed GPT-3.5 and Gemini Pro on both high and low-resource languages and in some cases matched or even surpassed GPT-4 [13].

Culture embodies a nation's historical experience and core values, shaping distinctive perspectives and approaches to social issues. For example, Germany abolished capital punishment with the enactment of its Basic Law in 1949 and has since actively advocated for its global abolition—a stance rooted in lessons drawn from the Nazi regime's abuse of the death penalty. By contrast, Japan retains capital punishment but rarely carries out executions, and public opinion polls indicate majority support for its continued existence. These contrasts illustrate how historical and cultural differences give rise to divergent positions on capital punishment.

As global hubs of technological innovation, China and the United States currently lead the world in the development of LLMs. However, their differing social values and cultural backgrounds may lead to systematic differences in LLM behavior and outputs. For instance, Li Jun et al. [14] found that China's education system emphasizes rote memorization and fosters respect for authority figures and traditional knowledge, whereas the U.S. system tends to encourage skepticism toward authority. Therefore, we propose evaluating Chinese cultural knowledge in LLMs developed in China and in the United States to assess potential differences. Since the emergence of ChatGPT, LLMs have entered a phase of rapid development. Currently influential global LLMs include GPT-5.1, DeepSeek-V3.2, and Gemini Pro, whose developers are primarily based in China and the United States. In this study, we select ten representative models from these developers for systematic testing. This study not only reveals differences in how LLMs from different countries respond to questions about Chinese culture but also provides a useful reference for future research on LLMs in the cultural domain.

## 2 Related Work

### 2.1 Large Language Models (LLMs)

This study focuses on globally influential and representative large language models (LLMs) that were selected to undergo testing on Chinese cultural questions. Given the rapid pace of development in the LLM field, rankings based on different evaluation dimensions vary substantially. In the November general ranking of the SuperCLUE Chinese LLM evaluation benchmark [15], GPT-5.1 ranked first, followed by Claude Sonnet 4.5, DeepSeek-V3.2, Gemini 2.5 Pro, and Kimi-K2. In August, DBC Consulting released a Top 30 ranking [16] based on technical capability and commercial implementation, in which ByteDance's Doubao model and Alibaba's

Qianwen series occupied the top three positions, followed closely by Wenxin. Notably, in November, Ant Group launched Lingguang, a multimodal general-purpose AI assistant capable of generating outputs such as 3D digital models and animations.

Building on the above, this study evaluates ten large language models (LLMs)—Lingguang, Kimi-K2, Claude Sonnet 4.5, Gemini 2.5Pro, Grok 4, GPT-5.1, DeepSeek-V3.2, Qwen3-Max, Doubao, and Wenxin Yiyan 5.0-Preview—using questions on traditional Chinese cultural knowledge. Among these models, Lingguang, Kimi-K2, DeepSeek-V3.2, Qwen3-Max, Doubao, and Wenxin-Yiyan are developed in China, whereas the remaining four are developed in the United States.

**2.2 Scoring Criteria**

For scoring, objective items such as multiple-choice and fill-in-the-blank questions are evaluated by directly comparing the model's answer with the reference answer. However, owing to the subjective nature of short-answer responses, when evaluating model answers against the reference answer, we consider a response correct if it differs only in wording but is consistent with the core content of the reference answer. By contrast, if the model's response is fundamentally inconsistent with the reference answer, it is judged as incorrect. For grading short-answer questions, we employ a multi-evaluator scoring system. Initially, two researchers specializing in Chinese language and literature assess each response. If both evaluators agree on the same grade, that grade is adopted. If disagreement occurs, the question is submitted to a third researcher for evaluation. The final grade is determined solely by the third researcher's assessment. Scoring examples can be found in Appendix.

Additionally, for short-answer questions with non-unique solutions—for example, Question 3, which requires the recitation of any three poems about the Winter Solstice—the reference answers include Du Fu's "Winter Solstice" and "Minor Solstice" and Bai Juyi's "Winter Solstice Night in Handan." If the model's response differs from these examples, we further verify whether the answer is in fact related to the Winter Solstice. For instance, GPT-5.1 responded to this question with Han Hong's "Cold Food Festival" and Liu Zongyuan's "River Snow," which do not satisfy the requirement of being Winter Solstice poems and therefore are marked as incorrect.

## 3 Experiment and Analysis

The core objective of this study is to examine whether large language models (LLMs) from China and the United States exhibit systematic differences in their responses to questions about Chinese culture. To ensure the accuracy and validity of the experimental data, we select relatively recent and influential LLMs from both countries. The following sections introduce the datasets and briefly describe the experimental procedures and results.

**3.1 Model Configuration**

All models were tested at the same time. Considering that online search involves data uploads and tests the model's existing knowledge base of Chinese culture, online search and deep thinking were disabled for all models. Additionally, since many models employ internal dynamic tuning for parameters such as inference temperature (Temperature) and Top_p without providing adjustment methods, all models were tested using their default parameters.

**3.2 Datasets**

The test questions used in this study were sourced from GitHub and comprise 456 multiple-choice questions, 144 fill-in-the-blank questions, and 29 short-answer questions. Data is sourced from[17].

Before finalizing all data, we conducted a data cleaning and filtering process. This primarily involved checking for duplicate questions, such as those with identical stems but different answer choice orders. However, some questions with the same theme but distinct content were retained. For example, Question 148 in Question Bank 1 and Question 65 in Question Bank 2 both relate to the Dragon Boat Festival, yet they assess different specific aspects. Therefore, both questions were kept. After completing the cleaning of all questions, we standardized the question numbers and formatting issues.

The questions span a broad range of content and are categorized into six major sections: traditional cultural knowledge (278 questions), literary classics (98 questions), history (89 questions), opera and music (52 questions), poetry and prose (68 questions), and notable historical figures (44 questions). These categories encompass fields such as general history, linguistic knowledge, and the history of Chinese literature.

### 3.3 Experimental Procedure

This study employs a direct-input procedure to pose questions to large language models (LLMs). For models such as Lingguang that cannot process entire question sets at once, the questions are entered in segments. Before testing each model, we provide an instruction prompt such as: "I will ask you questions related to Chinese culture. Please provide the correct answer." Responses from all LLMs are scored at 1 point per question, the scores are averaged by country, and the resulting error distributions are compared between Chinese and U.S. models.

### 3.4 Results

Figure 1 summarizes the performance of LLMs on multiple-choice, fill-in-the-blank, and short-answer questions. In this bar chart, the blue, orange, and gray bars represent the accuracy of each LLM on the multiple-choice, fill-in-the-blank, and short-answer sections, respectively. The figure reveals a consistent trend across most LLMs: the highest accuracy is achieved on multiple-choice questions, followed by fill-in-the-blank questions, with short-answer questions yielding the lowest accuracy.

For multiple-choice questions, most models achieved high accuracy, generally above 90%, with the exception of GPT-5.1 (89.69% accuracy). For fill-in-the-blank questions, accuracy varied substantially: some models reached 100% accuracy (e.g., Doubao, Qwen3-Max), whereas Grok4 recorded a low of 75.69%. For short-answer questions, all models performed worse on fill-in-the-blank and multiple-choice questions. Kimi-K2 achieved the highest accuracy at 82.76%, while GPT-5.1 recorded the lowest score at 20.69%.

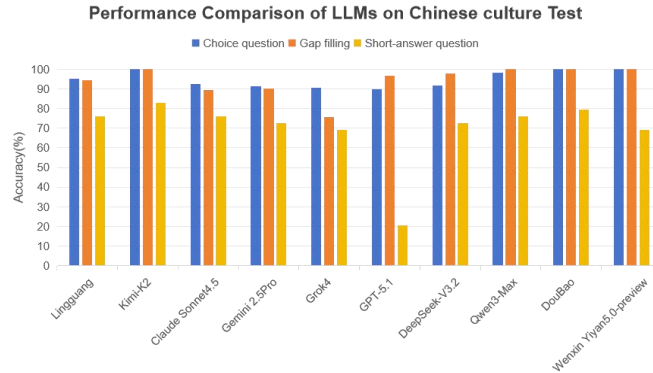

**Figure 1 Scoring by question type: Each model was tested on 456 multiple-choice questions, 144 fill-in-the-blank questions, and 29 short-answer questions. The accuracy rate for each type is calculated by dividing the number of questions answered correctly by the total number of questions in that type.**

Notably, the Chinese LLM Kimi-K2 demonstrated exceptional performance in this Chinese culture assessment, achieving 100% accuracy on both multiple-choice and fill-in-the-blank questions and the short-answer questions achieved an 82.76% accuracy rate. Among large language models from the United States, Claude Sonnet 4.5 demonstrated relatively stable performance across all question types in this test compared to other models from the same country. In contrast, Grok4 and GPT-5.1 showed lower accuracy rates in fill-in-the-blank and short-answer questions, respectively. Chinese large language models (LLMs) achieved 100% accuracy on multiple-choice and fill-in-the-blank questions, while no American LLM attained perfect scores in these categories.

Next, we calculated the average accuracy for each question type by country of origin (see Figure 2). Overall, Chinese LLMs demonstrated higher accuracy than their U.S. counterparts when answering questions about

Chinese culture, with average accuracy consistently exceeding 90%. Among the three question types, the accuracy gap between large language models (LLMs) from different countries in multiple-choice and fill-in-the-blank questions—6.52% and 10.71% respectively—is bigger than the gap in short-answer questions (16.38%). This indicates that large language models (LLMs) demonstrate greater proficiency in addressing highly structured questions with definitive answers. However, for short-answer questions involving stronger subjectivity, both types of models achieved accuracy rates below 80%, with a 16.38% disparity between them. This suggests that LLMs still exhibit slight shortcomings when responding to more subjective inquiries, with Chinese-based LLMs performing better than their American counterparts.

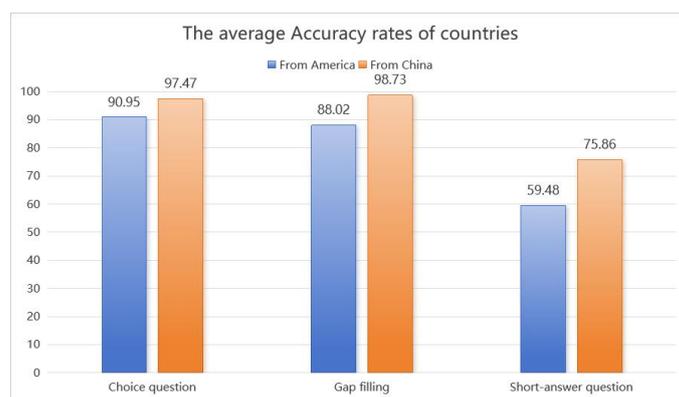

**Figure 2 The average accuracy rates of countries**

The above analysis characterizes the performance of Chinese and U.S. LLMs by question type. Next, we analyze performance by question category, where the error rate is calculated by first tallying the total number of errors made by a country's LLMs in that category and then averaging across models. The results of this analysis are shown in Figure 3. Regardless of whether performance is broken down by question type or by subject category, U.S. LLMs overall performed slightly worse than their Chinese counterparts in this test of traditional Chinese culture. Figure 3 shows that U.S. LLMs performed worst when answering questions related to literary classics, poetry, and famous figures. For the other three categories, their error rates hovered around 7%, showing only minimal differences relative to Chinese LLMs.

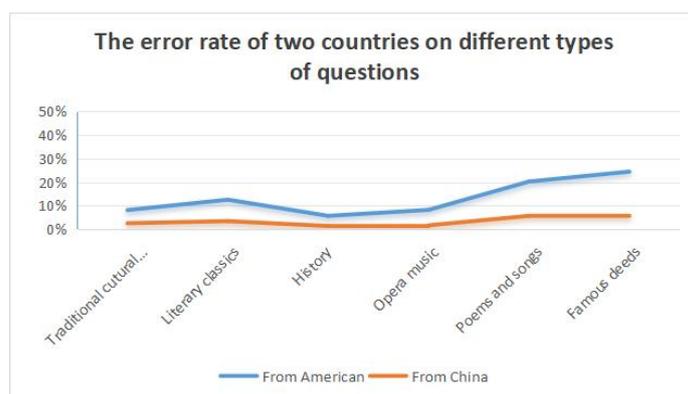

**Figure 3 The error rate of two countries on different types of questions**

It is noteworthy that Chinese LLMs exhibited their highest error rate when answering questions about famous historical figures, with error rates in this category exceeding 5%, higher than in any other category. Both Chinese and U.S. LLMs performed well when answering questions about history and opera music. Among the six categories of traditional Chinese culture questions, both Chinese and U.S. LLMs exhibited their highest error rates in the categories of poetry and historical figures. For instance, for the 53 fill-in-the-blank questions about poetry, the error rate reached 50%. When asked to identify the historical figure described in the line "若与功名论，几与卫霍同" (If measured by military achievements, he rivals Wei Qing and Huo Qubing), five out of ten LLMs provided incorrect answers: DeepSeek-V3.2, Lingguang, Gemini 2.5 Pro, Grok 4, and Claude Sonnet 4.5. The error rate for Question 24 on the Four Great Tragedies of the Yuan Dynasty reached a staggering 70%. Large language models that provided incorrect answers included GPT-5.1, Claude Sonnet 4.5, Kimi-K2, Gemini 2.5 Pro, DeepSeek-V3.2, Grok4, and Wenxin Yiyan 5.0-preview.

## 4 Discussion

In an era of rapid globalization and technological progress, cross-cultural communication is indispensable, particularly for countries at the center of global economic and technological development. However, cultural differences continue to pose substantial challenges for cross-cultural exchanges between China and the United States, as the two countries often regard issues from divergent perspectives.

In this study, we evaluated representative LLMs developed in China and in the United States on questions about traditional Chinese culture. The results indicate that U.S. LLMs performed worse than their Chinese counterparts on these traditional Chinese cultural questions, with the largest performance gaps observed for fill-in-the-blank items and for questions concerning the deeds of famous Chinese figures. This, in turn, underscores the importance of developing localized LLMs that can better understand and help preserve the history and culture of specific nations and linguistic communities. We hope that this research will stimulate further studies on the performance of LLMs in cultural contexts, and provide new perspectives for the culturally informed development of future LLMs.

**Appendix**

（1）　In response to short-answer question 2 regarding the origin of the bagu essay, Grok4 stated, "Derived from the classical essay style of the Song Dynasty, it was formally standardized as the required format for imperial examinations during the Ming Dynasty. Promoted by Zhu Yuanzhang to regulate scholars' thinking." The correct answer was "a writing style mandated by the Ming and Qing imperial examination system," but it did not explicitly mention the origin of the bagu essay. Consequently, one researcher considered Grok4's response correct. However, another researcher deemed it incorrect, citing the statement "promoted by Zhu Yuanzhang" as factually inaccurate. As the researchers were divided, the case was escalated to a third researcher, who also concluded that the statement contradicted historical records. Ultimately, Grok4's response was judged incorrect.

（2）　When addressing Short Answer Question 8 on the origin of the "Imperial Calendar," Claude Sonnet 4.5 responded with "Compiled by the Imperial Astronomical Bureau, reviewed by the emperor, and promulgated nationwide to represent the mandate of heaven." Both researchers deemed this answer excessively concise, lacking detailed explanation of the calendar's origin, and unanimously ruled it incorrect.